\documentclass[lettersize,journal]{IEEEtran}
\usepackage{caption}
\usepackage{amsmath,amsfonts}
\usepackage{algorithm}
\usepackage{tabularray}
\usepackage{color}
\usepackage{array}
\usepackage{textcomp}
\usepackage{stfloats}
\usepackage{float}
\usepackage{graphicx}
\usepackage{cite}
\usepackage{subfig}
\usepackage{caption}
\usepackage{subcaption}
\usepackage{makecell}
\usepackage{booktabs}
\usepackage{multirow}
\usepackage{algorithmicx}
\usepackage[noend]{algpseudocode}
\usepackage[table,xcdraw]{xcolor}
\usepackage[table,xcdraw]{xcolor}
\usepackage{color}
\usepackage{lineno}
\usepackage{nomencl}
\usepackage{amssymb}
\usepackage{url}

\usepackage{hyperref} 

\usepackage{arydshln}

\usepackage{amsthm}   
\theoremstyle{plain}

\usepackage{cleveref}
\crefname{table}{Table}{Tables}
\crefname{assumption}{Assumption}{Assumptions}
\crefname{section}{Section}{Sections}
\crefname{appendix}{Appendix}{Appendixes}
\crefname{algorithm}{Algorithm}{Algorithms}
\crefname{figure}{Figure}{Figures}
\crefname{theorem}{Theorem}{Theorems}
\crefname{lemma}{Lemma}{Lemmas}
\crefname{definition}{Definition}{Definitions}
\renewcommand{\eqref}[1]{Eq.~(\textup{\ref{#1}})}

\usepackage{bm} 

\usepackage{algorithm}
\usepackage{algorithmicx}
\usepackage{algpseudocode}
\usepackage{amsthm}   
\theoremstyle{plain}

\usepackage{xcolor}
\usepackage{todonotes}
\usepackage{pifont}

\usepackage{cleveref}
\crefname{table}{Table}{Tables}
\crefname{assumption}{Assumption}{Assumptions}
\crefname{section}{Section}{Sections}
\crefname{appendix}{Appendix}{Appendixes}
\crefname{algorithm}{Algorithm}{Algorithms}
\crefname{figure}{Figure}{Figures}
\crefname{theorem}{Theorem}{Theorems}
\crefname{lemma}{Lemma}{Lemmas}
\crefname{definition}{Definition}{Definitions}
\renewcommand{\eqref}[1]{Eq.~(\textup{\ref{#1}})}
\usepackage{bm} 
\newcommand{\cmark}{\ding{51}}  
\newcommand{\xmark}{\ding{55}}  

\makenomenclature

\begin{document}
\graphicspath{ {./picture/} }
\bibliographystyle{IEEEtran}

\title{MSCA-Net: Multi-Scale Context Aggregation Network for Infrared Small Target Detection}
\author{  Xiaojin Lu, Taoran Yue, Jiaxi Cai, Yuanping Chen, Cuihong Lv, Shibing Chu$^{*}$ 
\thanks{$^{*}$Corresponding author.}      
\thanks{The authors are with School of Physics and Electronic Engineering, Jiangsu University, China.(Corresponding author e-mail:c@ujs.edu.cn)}}


\maketitle

\begin{abstract} In complex environments, detecting tiny infrared targets has always been challenging because of the low contrast and high noise levels inherent in infrared images. These factors often lead to the loss of crucial details during feature extraction. Moreover, existing detection methods have limitations in adequately integrating global and local information, which constrains the efficiency and accuracy of infrared small target detection. To address these challenges, this paper proposes a network architecture named MSCA-Net, which integrates three key components: Multi-Scale Enhanced Dilated Attention mechanism (MSEDA), Positional Convolutional Block Attention Module (PCBAM), and Channel Aggregation Feature Fusion Block (CAB). Specifically, MSEDA employs a multi-scale feature fusion attention mechanism to adaptively aggregate information across different scales, enriching feature representation. PCBAM captures the correlation between global and local features through a correlation matrix-based strategy, enabling deep feature interaction. Moreover, CAB enhances the representation of critical features by assigning greater weights to them, integrating both low-level and high-level information, and thereby improving the model’s detection performance in complex backgrounds. The experimental results demonstrate that MSCA-Net achieves strong small target detection performance in complex backgrounds. Specifically, it attains mIoU scores of 78.43\%, 94.56\%, and 67.08\% on the NUAA-SIRST, NUDT-SIRST, and IRTSD-1K datasets, respectively, underscoring its effectiveness and strong potential for real-world applications.
\end{abstract}


\maketitle

\section{Introduction}
In multi-field application scenarios, infrared small target detection(IRSTD) plays an irreplaceable role, including military early warning \cite{bertrand2020infrared}, disaster early warning \cite{zhang2010applications}, secure night surveillance \cite{deng2016small}, traffic monitoring \cite{sun2020infrared}, aerospace imaging \cite{karim2013infrared} and agriculture \cite{awais2022comparative}. In the military domain, it facilitates the rapid identification of enemy targets, enhances battlefield situational awareness, and enables the early detection of potential threats, thereby improving the accuracy and efficiency of strategic decision-making \cite{xia_ultra_2025}. In disaster rescue operations, IRSTD precisely locates trapped individuals, supporting swift search and rescue efforts, significantly enhancing rescue efficiency and reducing casualties \cite{fekete2015critical}. In environmental monitoring, this technology is employed for meteorological observations, atmospheric pollution detection, early warning of forest fires, and other applications. This allows for the timely detection of potential hazards and provides accurate data support to relevant agencies, ensuring that disaster response measures are both prompt and effective \cite{asadzadeh2022uav}. In the field of agriculture, IRSTD is equipped with unmanned aerial vehicles to acquire remote sensing data of rice and wheat, and combined with a neural network model, provides intelligent solutions for optimizing crop harvesting schemes and monitoring diseases and pests \cite{cong2022research,zhu2022using}. Moreover, owing to its passive operation, robust target recognition capabilities, and all-weather functionality, IRSTD also holds significant value in small target detection.
\newline \indent However, despite its considerable application potential, IRSTD still faces numerous practical challenges.In long-distance imaging application scenarios, weak and small targets often present pixel-level representations—specifically manifested as contrast characteristics weaker than 0.15, signal-to-noise ratio levels less than 1.5, and pixel proportions not exceeding 0.15\% \cite{chapple1999target}—resulting in a lack of distinct visual features such as color and texture, which makes segmentation particularly challenging in complex backgrounds. Furthermore, under the interference of complex, dynamically changing backgrounds and noise from wave and cloud effects, these small targets are often prone to blurring or occlusion, further complicating detection. Consequently, while effectively reducing background noise, enhancing the model's ability in feature extraction and achieving multi-scale detection of small targets remains a challenging task.
\newline \indent To solve these problems, we proposed the MSCA-Net model and designed the Multi-Scale Enhanced Dilated Attention mechanism (MSEDA). By constructing a bidirectional feature interaction channel, the adaptive enhancement of multi-scale features and the deep coupling of local-global semantics were achieved. Secondly, the Positional Convolutional Block Attention Module (PCBAM) \cite{pramanik2024dau} is introduced, and the spatial - channel - positional triple attention mechanism is adopted to enhance the feature responses of the key regions. Finally, this study embedded the Channel Aggregation Feature Fusion Block (CAB) \cite{li2022moganet}. By assigning greater weighting of the critical information channels, the fusion of key features is effectively enhanced. These three modules form a collaborative optimization system, solving the problem of feature discrimination under dynamic background interference and providing a theoretical solution for precise positioning and segmentation in highly dynamic and complex backgrounds.
\newline \indent Experiments show that the multi-scale semantic enhancement network constructed in this study demonstrates significant advantages in the segmentation accuracy and detection efficiency of infrared dim and small targets, providing theoretical support and practical guidance for the innovative application of deep learning technology in the field of infrared imaging. Specifically, the contributions of this research are reflected in the following three aspects:
\newline \indent(1) This study constructs a deep learning architecture for infrared dim and small target detection—MSCA-Net, based on cross-scale feature collaboration and context-aware modeling. This network adaptively extracts critical features across different scales while facilitating cross-layer feature fusion, significantly improving target detection performance and segmentation accuracy in complex backgrounds.
\newline \indent(2) We design a Multi-Scale Enhanced Dilated Attention Module (MSEDA), which employs a multi-scale feature fusion attention mechanism to adaptively aggregate information across different scales. This mechanism significantly enhances the robustness of the model in environmental perception in complex scenarios by dynamically strengthening the representation of key semantics, and ultimately achieves a breakthrough in pixel-level detection accuracy.
\newline \indent(3) We introduce the PCBAM module and design the Channel Aggregation Feature Fusion Block(CAB). PCBAM combines the Position Attention Module(PAM) and the cross-space channel interaction mechanism(CBAM) to achieve cross-layer feature fusion effectively, thereby enhancing the model’s focus on key regions. Moreover, CAB facilitates contextual information interaction by aggregating and weighting critical channels, facilitating efficient transmission of beneficial features and further improving the model’s target detection capability in complex backgrounds.
\newline \indent The remainder of this paper is organized as follows. Section \hyperref[sec:related]{2} reviews related previous work. Section \hyperref[sec:method]{3} details the specifics of our detection network. Section \hyperref[sec:experiments]{4} constructing the experimental verification system and demonstrating the effectiveness of the proposed algorithm through the method of quantitative division. Finally, Section \hyperref[sec:conclusion]{5} concludes the paper.

\section{Related work} 
\label{sec:related}
We can broadly classify existing IRSTD methods into two paradigms: model-driven approaches centered on physical feature modeling and data-driven approaches using deep learning \cite{zhang2024global}. This section briefly reviews the application of traditional model-driven methods and current data-driven methods in the segmentation of infrared small targets, focusing particularly on the significant research progress achieved by the U-Net architecture and its variants in this field.
\subsection{Model-driven approaches}    
\label{Traditional}
Over the past several decades, researchers have proposed a variety of model-driven IRSTD algorithms. These methods encompass techniques inspired by the human visual system (HVS) \cite{chen2013local,kim2012scale}, low-rank approximation approaches \cite{gao2013infrared,zhu2019infrared,zhang_global_2024}, top-hat filtering \cite{bai2010analysis}, and local contrast-based algorithms \cite{han2020infrared}. They typically rely on explicit mathematical models and assumptions, demonstrating excellent performance when there is a pronounced contrast between the target and its background. However, these traditional approaches exhibit certain limitations, especially when targets are embedded in complex backgrounds or when loud background noise is present. Specifically, when targets are interfered with by intricate backgrounds or high-contrast noise, the detection performance of conventional methods often deteriorates markedly, leading to a high rate of false alarms. This degradation arises because model-driven methods usually assume a distinct contrast difference between the target and the background—a condition that is frequently violated in real-world scenarios. Consequently, in dynamic and complex environments, particularly in practical applications such as military reconnaissance and security surveillance, these traditional methods often fail to achieve the required accuracy.

\subsection{Data-driven approaches}   
\label{Data-driven}
To overcome the limitations of conventional model-driven approaches, researchers have shifted their focus toward data-driven deep learning methods. Compared with traditional techniques, these methods exhibit a distinct advantage in suppressing background noise and have achieved promising results in IRSTD.
\newline \indent Researchers initially introduced convolutional neural networks (CNNs) \cite{krizhevsky2012imagenet} for IRSTD, significantly enhancing detection performance because of their powerful feature extraction capabilities. This breakthrough preliminarily validated the immense potential of data-driven models in this domain. However, the conventional CNN down-sampling process, while high-level features are extracted, also results in a gradual loss of spatial information, posing a substantial challenge for the precise localization of small targets. Moreover, Unlike simple target detection tasks, segmentation requires not only the identification of the target but also the accurate delineation of its boundaries, which is crucial for the effective recognition and tracking of infrared small targets. Owing to their lack of pixel-level prediction mechanisms, traditional CNNs models and standard classification networks are often inadequate in meeting the stringent requirements for precise contour extraction inherent in target segmentation tasks.
\newline \indent To overcome these issues, researchers have adopted the encoder–decoder U-Net architecture to address the shortcomings of conventional CNNs in small target detection. Originally introduced by Ronneberger et al. \cite{ronneberger2015u} for medical image segmentation, U-Net’s distinctive encoder–decoder design excels in precise segmentation tasks. By merging high-level feature maps with low-level ones through skip connections, the architecture effectively restores the spatial details of targets. This capability enables U-Net to accurately delineate small target regions in infrared images and maintain robust performance against complex backgrounds. Consequently, U-Net has gradually become the predominant approach in IRSTD, and its numerous variants (e.g., 3D-U-Net \cite{cciccek20163d} and Attention U-Net \cite{oktay2018attention}) have further enhanced both detection and segmentation performance. For example, in 2019, Wang et al. \cite{wang2019miss} constructed a generative adversarial network architecture called MDvsFA-cGAN, which is based on the GAN paradigm and aims to achieve accurate segmentation of small infrared targets. In 2021, Dai et al.\cite{dai2021asymmetric} designed an asymmetric context modulation architecture (ACM), which enhanced the feature interaction integrity of the bottom-up modulation channel by establishing an inverse feature compensation path, effectively solving the multi-scale semantic mismatch problem in traditional methods. In the same year, Dai et al. \cite{dai2021attentional} proposed ALCNet, which transforms the classic local contrast analysis method into a local attention-guided method. Next year, Zhang et al. \cite{zhang2022isnet} developed the edge block aggregator network ISNet, aiming to construct cross-level edge features and significantly reduce the context-target semantic gap.
\newline \indent Nevertheless, a standalone U-Net architecture still struggles with segmentation accuracy in IRSTD. To resolve this bottleneck, researchers have integrated the U-Net architecture with advanced attention mechanisms and introduced a cross-level feature collaboration strategy to enhance the model’s responsiveness to critical information regions, thereby further improving the accuracy of infrared small target recognition and segmentation. For example, Ren D et al. \cite{li2022dense} constructed DNANet, a dense nested attention network based on the U-Net++ architecture, which can realize an end-to-end network from degraded space to clear image space through topological connections. Similarly, Wu X et al. \cite{wu2022uiu} improved U-Net and constructed a recursive nested attention framework (UIU-Net), which recursively embeds the U-Net subnet into the backbone U-Net and combines channel and spatial attention for multi-level and multi-scale object representation learning. 
\newline \indent In addition, Sun et al. \cite{sun2023receptive} designed the RDIAN network, which combines receptive field and direction-aware attention mechanisms; it achieves feature capture of multi-scale receptive domains through convolution kernel configuration, while using multi-directional attention mechanisms to enhance the target information in the shallow feature space. Hou et al. \cite{Hou2022ISTDU-Net} introduced ISTDU-Net, an infrared small target detection network that converts single-frame infrared images into pixel-wise target probability maps. To strengthen the representation of small target features, the network incorporates feature map groups during down-sampling; fully connected layers, added within skip connections, effectively suppress background interference, improving target-background contrast. Xu et al.~\cite{yuan2024sctransnet} proposed the SCTransNet network, which introduces the core module SSCA (Spatial-Embedded Single-Head Channel Cross-Attention) to enhance semantic interactions across different feature levels and effectively model long-range contextual dependencies in images, thereby significantly improving the performance of infrared small target detection and segmentation accuracy.
\newline \indent In addition to U-Net-based architectures, other network models have also played pivotal roles in infrared small target detection. For example, Liu et al. \cite{liu_moumamba_2025}  constructed the multi-order U-shaped Mamba network, and the Multi-Order 2D-Selective Scan (MO-SS2D) module and Local-Guided 2D-Selective Scan (LG-SS2D) module were constructed. The collaborative improvement of detection accuracy and reasoning efficiency has been achieved through the dynamic scanning mechanism of hierarchical feature interaction and local context guidance. Ma et al. \cite{ma_wavelet_2025}  proposed integrating discrete Wavelet Transform (DWT) into a convolutional neural Network (CNN), and designed a hybrid attention mechanism, namely small waveguide direction Transformer (WST), to fully enhance the spatial and cross-channel characteristics of wavelet transform, achieving the collaborative optimization of detection accuracy and anti-noise performance. In addition, Yue et al. \cite{yue_yolomst_2025a} presented the YOLO-MST model, which improves detection accuracy by incorporating super-resolution technology and multi-scale feature extraction, achieving more efficient infrared target detection while enhancing overall performance.
\newline \indent In summary, integrating U-Net with a multi-scale attention mechanism can enhance feature representation, thereby improving segmentation performance. Moreover, cross-layer feature fusion is crucial for optimizing model performance, as it effectively integrates semantic information across different layers and strengthens the perception of infrared small targets. Therefore, exploring an effective approach to integrate the U-Net architecture, multi-scale attention mechanisms, and cross-layer feature fusion remains highly valuable for improving feature extraction accuracy and target perception capability.

\section{Methodology}
\label{sec:method}
\begin{figure*}[htbp]
	\centering
	\includegraphics[width=.98\textwidth]{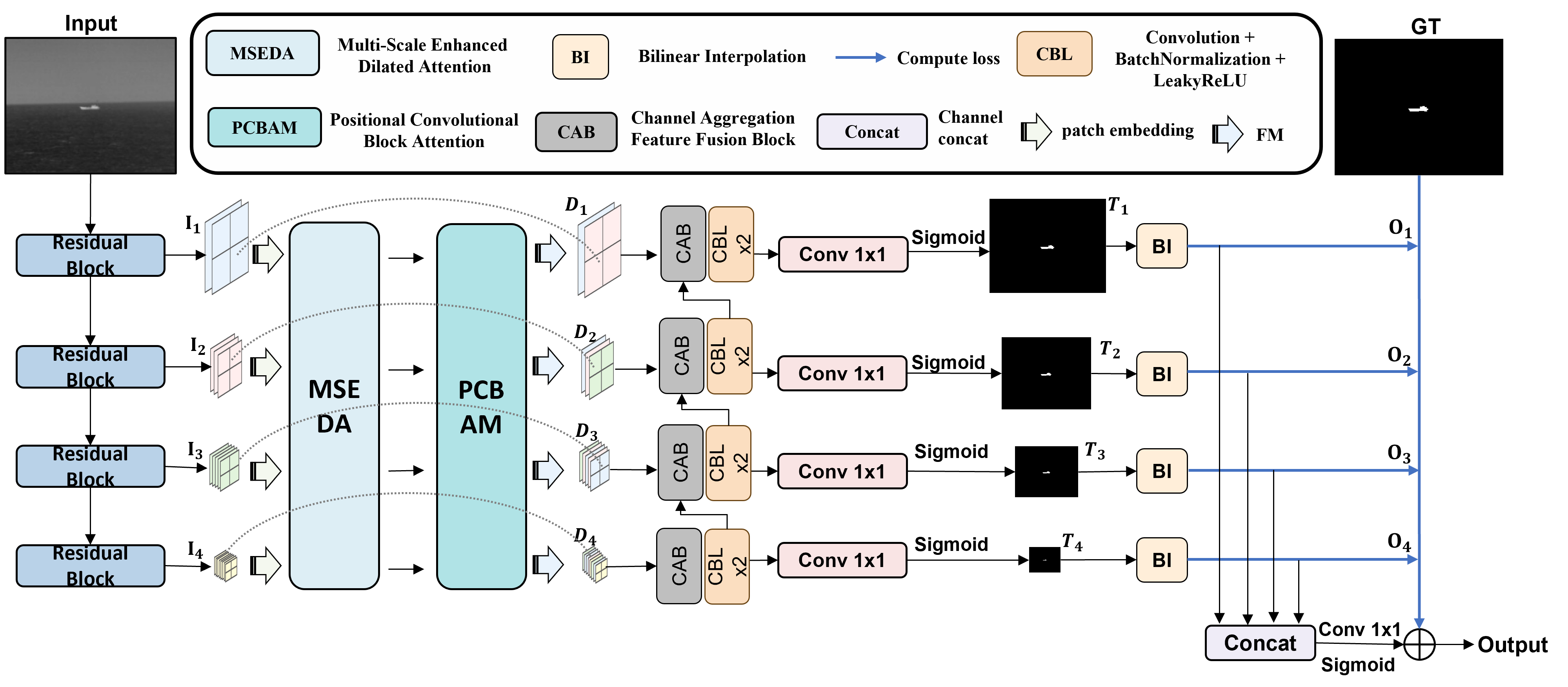}
	\caption{MSCA-Net overall architecture diagram.} 
	\label{fig:overall architecture}
\end{figure*}

Figure \ref{fig:overall architecture} presents the holistic structure of the MSCA-Net framework. The proposed model is built upon the U-Net baseline. Initially, four residual blocks (RBs) \cite{he2016deep} and max-pooling layers are employed to extract images features \( I_i \in \mathbb{R}^{C_i \times \frac{H}{i} \times \frac{W}{i}} \), (\( i = 1, 2, 3, 4 \)). The corresponding channel dimensions are \( C_1 = 32 \), \( C_2 = 64 \), \( C_3 = 128 \), and \( C_4 = 256 \). 
\newline \indent Next, patch embedding is applied to each \( I_i \) using convolutions with kernel sizes and strides of \( P \), \( P/2 \), \( P/4 \), and \( P/8 \), respectively. The resulting features are fed into the MSEDA and PCBAM modules for full-level semantic feature blending. Subsequently, to obtain \( D_i \in \mathbb{R}^{C_i \times \frac{H}{16} \times \frac{W}{16}} \), (\( i = 1, 2, 3, 4 \)), the features are then restored to the original encoder size through a feature mapping (FM) module \cite{yuan2024sctransnet}, which integrates bilinear interpolation, convolution, batch normalization, and ReLU activation. In parallel, a residual connection is introduced to effectively bridge the semantic gap between the encoder \( I_i \) and decoder \( D_i \) by fusing their respective features. These representations are fused via the CAB module, followed by a CBL operation (Convolution + Batch Normalization + LeakyReLU). A final convolutional layer and a Sigmoid activation function are used to produce the prediction maps \( T_i \), (\( i = 1, 2, 3, 4 \)). Bilinear interpolation is employed to restore the image size of the output. 
\newline \indent Finally, the Binary Cross-Entropy (BCE) loss \cite{wu2022uiu} is used to calculate the difference between the predicted outputs and the ground truth (GT). The detailed structures of MSEDA, PCBAM, and CAB are described in Sections~\hyperref[sec:MSEDA]{3.1}, \hyperref[sec:PCBAM]{3.2}, and \hyperref[sec:cab]{3.3}, respectively.

\subsection{Multi-scale enhanced dilated attention module}   
\label{sec:MSEDA}
\begin{figure*}[htbp]
	\centering
	\includegraphics[width=.98\textwidth]{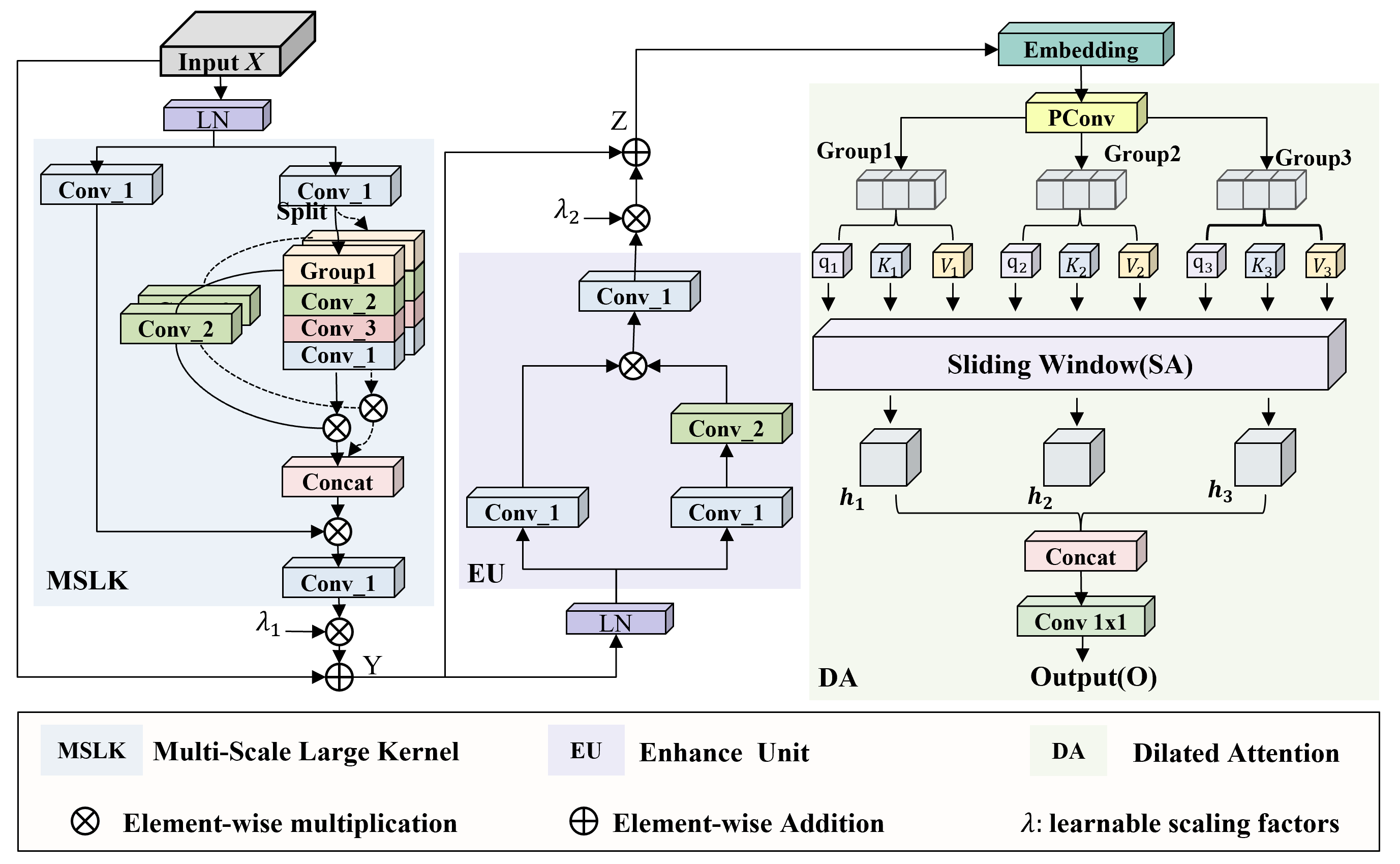}
	\caption{MSEDA module architecture.} 
	\label{fig:MSEDA}
\end{figure*}

Small targets in images often exhibit low contrast and subtle features, rendering conventional convolutional neural networks ineffective at extracting their information. To address this challenge, we drew inspiration from the transformer paradigm to redesign the basic convolutional block for feature extraction. Building on this foundation, we incorporated the core concepts of MAB \cite{wang2024multi} and MSDA \cite{jiao2023dilateformer} into the proposed MSEDA module, specifically optimizing it for the characteristics of infrared small targets. Leveraging hierarchical feature interactions alongside an adaptive weight allocation mechanism effectively enhances both the representation and fusion efficiency of cross-scale features. This approach substantially improves the model’s accuracy in capturing complex pattern features, offering a novel architectural optimization for multi-scale feature extraction tasks. As shown in Figure \ref{fig:MSEDA}, the MSEDA module employs a hybrid strategy that integrates multi-scale large kernel attention embedding with dilated convolutions, further increasing its ability to capture multi-scale information.
\newline \indent Given a feature map $X$, we apply layer normalization (LN) to standardize the features, enhancing training stability and accelerating model convergence. After normalization, we employ three convolutions with distinct parameter settings to extract features at specific scales or orientations. The first convolution uses a small kernel to capture local fine-grained details, whereas the second adopts a larger kernel to focus on global contextual information. The third operation incorporates dilated convolutions with varying dilation rates to enlarge the receptive field, thereby balancing global and local feature representations. 
\newline \indent The extracted features from these three convolutional branches are subsequently fused and fed into an embedding layer to enrich feature diversity and semantic representation. The output feature map $X$ is then processed through additional convolutions to generate the corresponding $q$, $k$, and $v$ (query, key, and value), with feature channels split across three attention heads. Self-attention is applied within each group, while different dilation rates are used to refine feature representations across groups. The aggregated features from all the groups are then concatenated to facilitate intergroup information exchange. Finally, a 1×1 convolution layer is employed to project the refined feature representations. Specifically, the MSEDA module is formally defined as follows:
\newline \indent For a given input feature $X$, the whole process of MSEDA can be formally described by the following equations (\ref{formula:mseda1}), (\ref{formula:mseda11}) and (\ref{formula:mseda2}):
\begin{equation}
	\begin{aligned}
		Y &= X + \lambda_1 \left(\mathrm{MSLK}\left({LN}(X)\right)\right) \\ 
		Z &= Y + \lambda_2 \left(\mathrm{EU}\left({LN}(Y)\right)\right) \\
		\label{formula:mseda1}
	\end{aligned}
\end{equation}
where $\mathrm{LN}(\cdot)$ and $\lambda_i$ are the layer normalization and learnable scaling factors, respectively. $\otimes$ denotes element-wise multiplication, and $\oplus$ denotes element-wise addition.
\newline \indent Subsequently, the processed features are passed through an embedding layer and then fed into the PConv module, which expands the channel dimension by a factor of three. The output is then reshaped and permuted to produce multiple groups of dilated heads, which are split into the corresponding query, key, and value tensors, denoted as \( Q_i, K_i, V_i \). Formally, this process can be expressed as:
\begin{equation}
	Q_i, K_i, V_i = \text{split}\left(\text{res}\left(\text{PConv}(\text{Eb}(Z))\right)\right), \, 1 \leq i \leq 3 \\ \\
	\label{formula:mseda11}
\end{equation}
here, \(\text{res}(\cdot)\) denotes the reshape operation and \(\text{Eb}(\cdot)\) represents the embedding function. 
\newline \indent These components are subsequently passed into the Dilated Attention (DA) module for further refinement, as formulated in equation~(\ref{formula:mseda2}):
\begin{equation}
	\small
	\begin{aligned}
		h_i &= \text{SA}(Q_i, K_i, V_i, r_i), \, 1 \leq i \leq 3 \\
		O &= \text{Conv}_{1 \times 1}(\text{Concat}[h_1, \ldots, h_n])
		\label{formula:mseda2}
	\end{aligned}
\end{equation}
where \( r_i \) is the expansion rate of the \( i \)-th head, \( Q_i \), \( K_i \), and \( V_i \) each represent a slice of the feature map fed to the \( i \)-th head. The outputs of all heads, \( \{ h_i \}_{i=1}^{n} \), are spliced together and then fed into the convolution layer for feature aggregation. At the default setting, we use a \( 3 \times 3 \) convolution kernel with expansion rates \( r = 1, 2, 3 \), and the receptive field sizes of different attention heads are \( 3 \times 3 \), \( 5 \times 5 \), and \( 7 \times 7 \), respectively.

\subsection{Positional convolutional block attention}   
\label{sec:PCBAM}
\begin{figure*}[htbp]
	\centering
	\includegraphics[width=.98\textwidth]{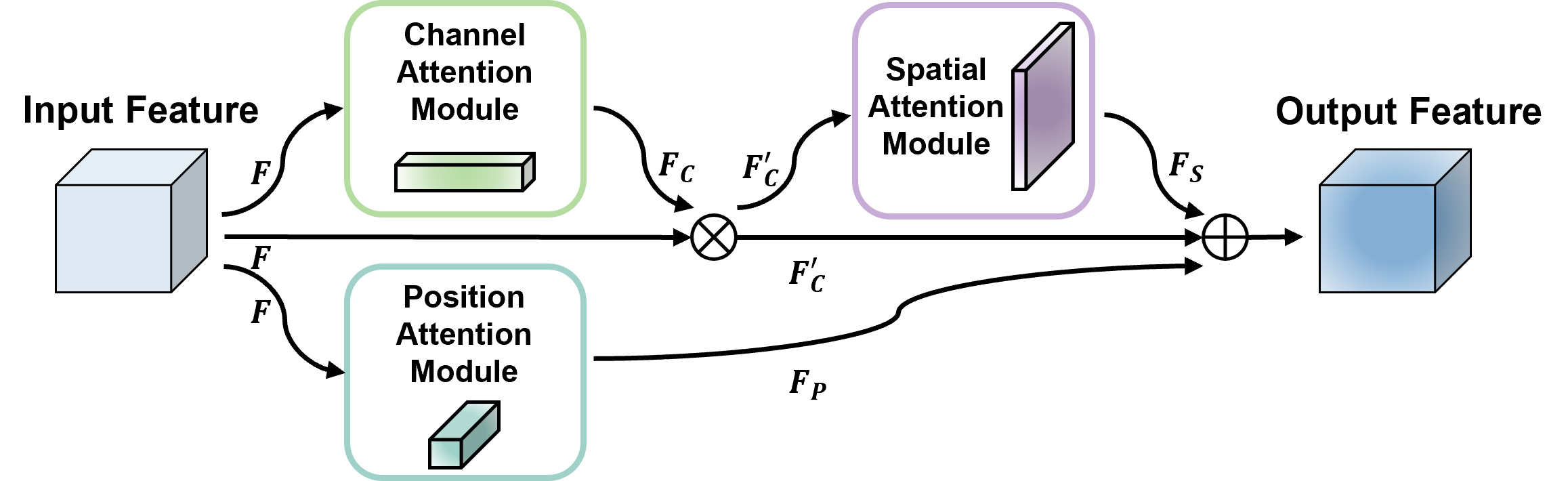}
	\caption{PCBAM module structure diagram.} 
	\label{fig:pcbam}
\end{figure*}

Considering the intrinsic constraints of current small-target detection approaches in integrating global semantics and achieving cross-layer feature fusion, this paper introduces the PCBAM module. Its design is inspired by the cross-modal interaction paradigm proposed in \cite{pramanik2024dau} and has been optimized to better suit infrared small-target detection tasks. PCBAM deeply integrates the position attention mechanism (PAM) with the channel-spatial attention module CBAM \cite{woo2018cbam} through an adaptive recalibration strategy and a dual-branch parallel architecture. By employing a dynamic weight allocation mechanism, PCBAM ensures the precise alignment of multi-granularity features, establishing global contextual dependencies along the channel dimension while leveraging spatial attention to guide geometric associations among local features. This ultimately forms a parameter-adaptive feature enhancement pathway. The experimental findings indicate that PCBAM provides notable improvements in refining feature weight allocation and promoting comprehensive interactions between holistic context and fine-grained details. To further explore the specific operational principles of the PCBAM module, we next provide a detailed mathematical analysis of its processing of the input feature $F$.
\newline \indent Given an input feature map $F$ with dimensions \( \text{C} \times \text{H} \times \text{W} \), where $\text{C}$, $\text{H}$, and $\text{W}$ denote the number of channels, height, and width, respectively. The PCBAM process is as follows. Initially, $F$ undergoes transformation within the channel attention module (CAM) to obtain \( F_C \):
\begin{equation}
	F_C = \sigma(\text{mlp}(\text{gap}(F)) + \text{mlp}(\text{gmp}(F)))
	\label{formula:pcbam1}
\end{equation}
in equation (\ref{formula:pcbam1}), \( \sigma \) denotes the sigmoid activation function, whereas \text{gap} and \text{gmp} denote the global average pooling  and global max pooling operations, respectively. The term \text{mlp} refers to a multilayer perceptron composed of two fully connected layers, where the first dense layer contains \( C \)-th channel units, and the second dense layer consists of \( \frac{C}{8} \)-th channel units. Subsequently, \( F_C' = F_C \otimes F \) is passed into the spatial attention module (SAM), where \( \otimes \) represents element-wise matrix multiplication. After processing through the SAM layer, \( F_C' \) is transformed into \( F_S \):
\begin{equation}
	F_S = f^{7 \times 7} \left[ \text{DL}(\text{gap}(F_C')) ; \text{DL}(\text{gmp}(F_C')) \right]
	\label{formula:pcbam2}
\end{equation}
in equation (\ref{formula:pcbam2}), \( f^{7 \times 7} \) indicates a convolutional operation with a \( 7 \times 7 \) kernel and a dilation rate of 4. DL refers to a dense layer, whereas \( ; \) represents the concatenation operation.
\newline \indent Next, we discuss the positional attention module. Given an input feature map \( {F} \in {R}^{H \times W \times C} \), the processing steps within this module are as follows. The input is processed by a convolutional layer, generating three new feature maps \( {B} \), \( {Z} \), and \({D} \), each with a dimensionality of \( {R}^{H \times W \times C} \). These feature maps are then reshaped into \( {R}^{N \times C} \) (where \( N = H \times W \)). Next, matrix multiplication is conducted between the transpose of \( {Z} \) and \( {B} \), and a softmax function is applied to yield the spatial attention map \( {S} \in {R}^{N \times N} \). \( {D} \) is subsequently reshaped back to \( {R}^{H \times W \times C} \). Finally, an element-wise addition is applied between \( {F} \) and the processed output, yielding the output \( {F}_P \in {R}^{H \times W \times C} \). The entire computation is formulated as shown in equations (\ref{formula:pcbam3}) and (\ref{formula:pcbam4}):
\begin{equation}
	s_{ji} = \frac{\exp(B_i \cdot Z_j)}{\sum_{i=1}^N \exp(B_i \cdot Z_j)}
	\label{formula:pcbam3}
\end{equation}
\vspace{-3mm}
\begin{equation}
	F_{P_j} = \alpha \sum_{i=1}^N s_{ji} \cdot D_i + F_j
	\label{formula:pcbam4}
\end{equation}
the term \( s_{ji} \) quantifies the influence of position \( i \) on position \( j \), whereas \( \alpha \) represents a scaling parameter initialized to zero and progressively learned and adjusted during training. Ultimately, the complete formulation of the PCBAM module is given by the following expression, with its detailed architecture illustrated in Figure \ref{fig:pcbam}.
\begin{equation}
	{F}_{\text{PCBAM}} = {F}_C' + {F}_S + {F}_P.
\end{equation}
\vspace{-3mm}

\subsection{Channel aggregation feature fusion block} 
\label{sec:cab}
\begin{figure*}[htbp]
	\centering
	\includegraphics[width=.95\textwidth]{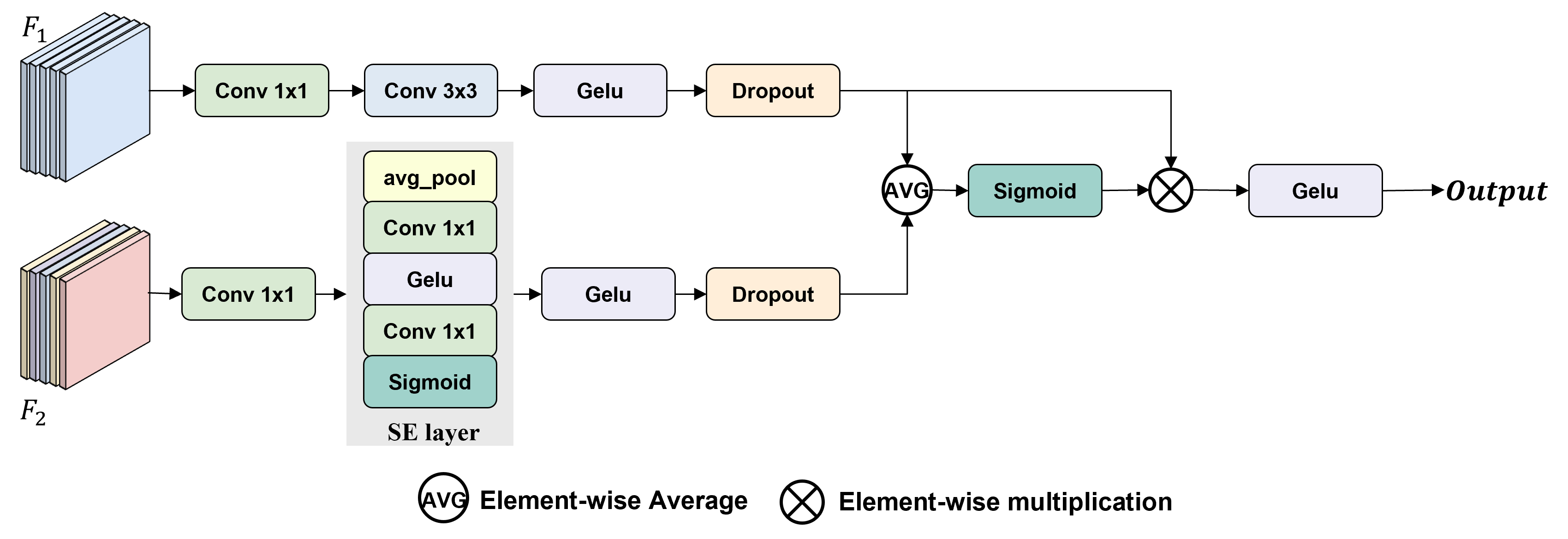}
	\caption{CAB structure diagram.} 
	\label{fig:cab}
\end{figure*}

Existing small-target detection models often struggle with inadequate target feature representation, particularly in complex backgrounds or multi-scale scenarios, where background information overwhelms or diminishes target features. To tackle this issue, the Channel Aggregation Feature Fusion Block (CAB) \cite{li2022moganet} is introduced to effectively mitigate the limitations in feature representation. As illustrated in Figure \ref{fig:cab}, CAB functions primarily by fusing the feature information of low-dimensional and high-dimensional, thereby enhancing the expression of critical features while suppressing redundant features. Specifically, it strengthens channel-wise relationships by assigning higher weights to essential channels, improving the model’s sensitivity to small targets and its robustness against complex backgrounds. Beyond enhancing feature distinguishability, CAB facilitates deep multi-scale information fusion, providing more precise feature representations for small-target detection tasks. Additionally, during the features fuse process, CAB performs weighted processing on the key channels, directing the model’s focus toward salient target characteristics and improving infrared small-target segmentation accuracy. Moreover, it preserves global contextual information while refining local feature capture, ensuring more reliable support for the final segmentation results. To further explore the specific role of the CAB, the following section presents a mathematical analysis of the processes of the input feature $F_1$ and $F_2$.
\newline \indent For the input feature $F_1$, its transformation is defined by equation (\ref{formula:cab1}):
\begin{equation}
	\begin{aligned}
		F_1 &= \text{Gelu}\left( \text{Conv}_{3 \times 3} \left( \text{Conv}_{1 \times 1} \left( (F_1) \right) \right) \right) \\
		F_1 &=  \text{Dropout}(F_1) 
		\label{formula:cab1}
	\end{aligned}
\end{equation}
similarly, the operation applied to $F_2$, its transformation is defined by equation (\ref{formula:cab11}):
\begin{equation}
	\begin{aligned}
		\mathrm{SE} &= \sigma \left( \mathrm{Conv}_{1 \times 1}\left( \mathrm{Gelu}\left( \mathrm{Conv}_{1 \times 1}\left( \mathrm{pool}(F_1) \right) \right) \right) \right) \\
		F_2 &= \text{Gelu}\left( \text{SE} \left( \text{Conv}_{1 \times 1} \left( (F_1) \right) \right) \right) \\
		F_2 &=  \text{Dropout}(F_2)
		\label{formula:cab11}
	\end{aligned}
\end{equation}
specifically, $\mathrm{pool}$ performs average pooling, and $\sigma$ denotes the Sigmoid function. The SE module applies weighted processing to deep features. Then, $F_1$ and $F_2$ are fused by element-wise average (AVG), followed by element-wise multiplication between the Sigmoid and Dropout-processed $F_1$ and the averaged features. Finally, the output is obtained through a Gelu activation. As shown in equation (\ref{formula:cab2}):
\begin{equation}
	\mathrm{Output} = \mathrm{Gelu}\left(\sigma\left( \mathrm{AVG}(F_1, F_2) \right) \otimes F_1 \right).
	\label{formula:cab2}
\end{equation}

\section{Experimental results}
\label{sec:experiments} 
This section validates the efficacy of the proposed method through extensive experiments. Section \hyperref[sec:dataset]{4.1} introduces the public dataset used in this study and describes its partitioning strategy. Section \hyperref[sec:Implementation]{4.2} details the training and testing procedures, including hyperparameter settings and hardware configurations. Section \hyperref[sec:Evaluation]{4.3} presents the evaluation metrics used to ensure the scientific rigor and reliability of the comparative analysis. Section \hyperref[sec:comparison]{4.4} conducts a comprehensive comparison between the proposed method and other state-of-the-art (SOTA) infrared small target detection networks from both qualitative and quantitative perspectives to assess the performance advantages. Section \hyperref[sec:Ablation]{4.5} systematically analyzes the role and contribution of each module within the framework through ablation studies, further verifying their critical impact on overall performance enhancement. Section \hyperref[sec:Visual]{4.6} presents experimental results that intuitively illustrate the detection effectiveness of different methods, strongly supporting the experimental conclusions.
\subsection{Dataset preparation}   
\label{sec:dataset}
To assess the efficacy of our introduced framework in IRSTD, we conducted experiments on three public datasets: NUAA-SIRST \cite{dai2021asymmetric}, NUDT-SIRST \cite{li2022dense} and IRSTD-1K \cite{zhang2022isnet}. Specifically, NUAA-SIRST contains 427 infrared images, NUDT-SIRST includes 1,327 images, and IRSTD-1K contains 1,001 images. All three datasets were divided into the training and test sets at a 7:3 ratio. These datasets cover infrared images captured at short, medium, and 950nm wavelengths, covering a wide range of remote sensing backgrounds, including the sky, ground, buildings, and oceans. In addition, the dataset contains multiple target types, including military targets such as drones and ships, further enriching the diversity of test scenarios. At the same time, these datasets also contain many challenging scenes, such as low contrast and complex backgrounds, ensuring a comprehensive assessment of model detection capabilities.

\subsection{Implementation details}   
\label{sec:Implementation}
To systematically evaluate the performance of the proposed methods, we compared MSCA-Net with SOTA IRSTD methods. The experimental benchmark consists of two paradigms: three model-driven approaches, Top-Hat \cite{zhu2019infrared} based on morphological filtering, WSLCM \cite{han2020infrared} based on local contrast, and IPI \cite{gao2013infrared} based on low-rank sparse decomposition. In deep learning, we compared seven cutting-edge models: the asymmetric context modulation network (ACM) \cite{dai2021asymmetric}, the attentional local contrast network (ALCNet) \cite{dai2021attentional}, the receptive field and direction-induced attention network (RDIAN) \cite{sun2023receptive}, the improved U-Net architecture (ISTDU-Net) \cite{Hou2022ISTDU-Net}, the interior attention-aware network (IAANet) \cite{wang2022interior}, the dense nested attention network (DNA-Net) \cite{li2022dense}, and the nested U-Net network (UIU-Net) \cite{wu2022uiu}.
\newline \indent To maintain the rigor of the comparative analysis, we employed standardized experimental datasets. First, all deep learning models were retained based on the training data set used by MSCA-Net to ensure the consistency of data division and training samples; Secondly, in the model reasoning stage, the traditional algorithm strictly follows the parameter configuration suggested in the original text. For the method requiring threshold segmentation, we fixed the threshold parameters recommended in various literature to avoid the influence of artificial adjustment on the results. \url{https://github.com/XinyiYing/BasicIRSTD} provides open-source implementations of most techniques.
\newline \indent Regarding the experimental setup, all tests were performed on an Intel Xeon Platinum 8481 CPU running a Linux operating system with 24.0 GB of RAM. All deep learning models were developed using the \texttt{torch} framework. To expedite the training process, an NVIDIA GeForce GTX 4090 D GPU with CUDA 11.8 was employed. The Adam optimizer with a smoothing constant of 0.9 was employed to optimize the MSE loss function. The initial learning rate was set to \( 10^{-3} \), with a minimum value of \( 10^{-5} \). We trained the model for 1,000 epochs, automatically adjusting the learning rate using a cosine annealing strategy during training. The batch size is set to 16, and the IoU threshold for NMS is set to 0.5.

\subsection{Evaluation metrics}   
\label{sec:Evaluation}
Segmentation approaches predominantly utilize pixel-wise evaluation metrics, such as IoU, nIoU, $P_d$, and $F_a$. These measures focus on assessing the shape characteristics of the detected targets.
\newline \indent (1) IoU: IoU is a pixel-level evaluation metric designed to measure an algorithm’s ability to delineate target contours. It quantifies model performance by computing the ratio between the intersection and union of the predicted and ground truth regions. The definition is as follows:
\begin{equation}
	\text{IoU} = \frac{\sum_{i=1}^{N} \text{TP}[i]}{\sum_{i=1}^{N} \left( T[i] + P[i] - \text{TP}[i] \right)}
\end{equation}
\newline \indent (2) nIoU: nIoU is a normalized variant of IoU. Its main goal is to provide a more precise assessment of small target segmentation performance while reducing the impact of larger targets on the outcomes.
\begin{equation}
	\text{nIoU} = \frac{1}{N} \sum_{i=1}^{N} \frac{\text{TP}[i]}{T[i] + P[i] - \text{TP}[i]}
\end{equation}
\newline \indent (3) $P_d$: $P_d$ is defined as the proportion of accurately detected targets $N_{\text{pred}}$ to the total number of targets $N_{\text{all}}$, expressed as:
\begin{equation}
	P_d = \frac{N_{\text{pred}}}{N_{\text{all}}}
\end{equation}
\newline \indent (4) $F_a$ is defined as the fraction of falsely detected target pixels $N_{\text{false}}$ over the total number of pixels in the image $P_{\text{all}}$, expressed as:
\begin{equation}
	F_a = \frac{N_{\text{false}}}{P_{\text{all}}}
\end{equation}
\newline \indent In addition to the evaluation methods based on fixed thresholds, we also utilize the Receiver Operating Characteristic (ROC) curve for a comprehensive assessment of the model. The ROC curve illustrates how the true positive rate ($P_d$) changes as the false positive rate ($F_a$) varies.

\subsection{Detection performance comparison}   
\label{sec:comparison}
\begin{figure*}[htbp]
	\centering
	\includegraphics[width=.99\textwidth]{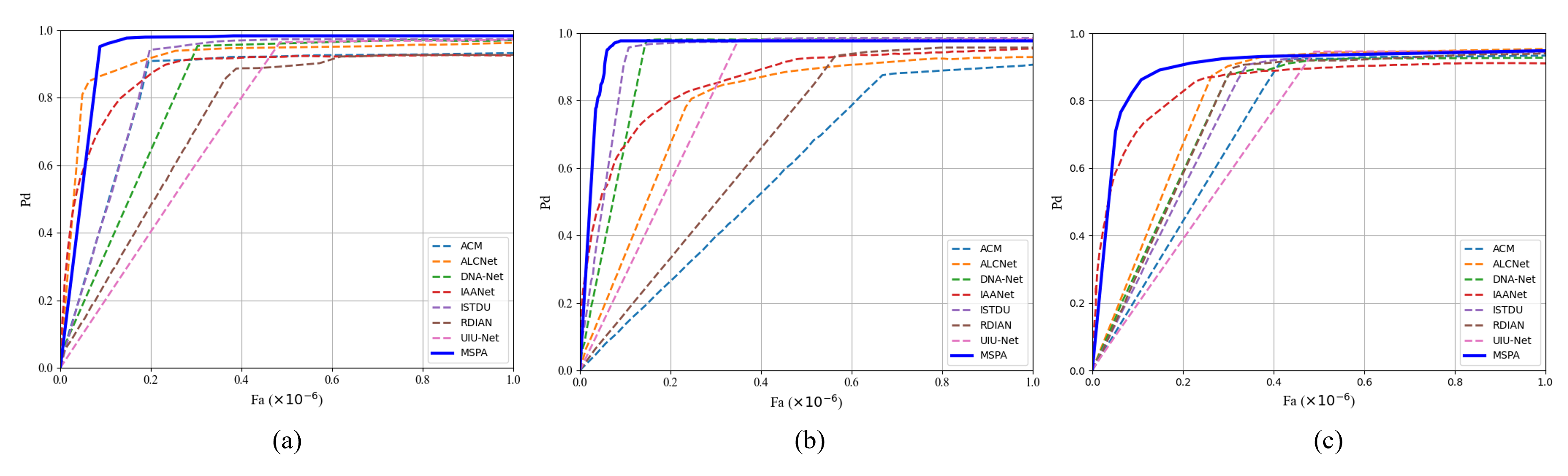}
	\caption{ROC curves of different methods. (a) NUAA-SIRST. (b) NUST-SIRST. (c) IRSTD-1K.} 
	\label{fig:ROC}
\end{figure*}

To demonstrate the effectiveness of our approach,we conducted a comparative study against several SOTA methods using three publicly available datasets: NUAA-SIRST, NUDT-SIRST, and IRSTD-1K. For these datasets, the mIoU reached 78.43\%, 94.56\%, and 67.08\%, respectively, whereas the nIoU reached 79.91\%, 94.36\%, and 67.15\%, respectively. The detailed quantitative results are summarized in Table~\ref{tab:comparison}.
\begin{table*}[t]
	\renewcommand{\arraystretch}{1.3}
	\centering
	\caption{Comparison with other SOTA methods on NUAA-SIRST, NUDT-SIRST and IRSTD-1K in $\mathrm{mIoU}~(\%)$, $\mathrm{nIoU}~(\%)$, $\mathrm{P}_\mathrm{d}~(\%)$, $\mathrm{F}_\mathrm{a}(10^{-6})$.}
	\label{tab:comparison}
	\setlength\tabcolsep{1.4mm}
	\begin{tabular}{c|cccc|cccc|cccc}
		\hline
		\multirow{2}{*}{Method} & \multicolumn{4}{c|}{NUAA-SIRST} & \multicolumn{4}{c|}{NUDT-SIRST} & \multicolumn{4}{c}{IRSTD-1K} \\
		\cline{2-13} 
		& mIoU & nIoU & $P_d$ & $F_a$ & mIoU & nIoU & $P_d$ & $F_a$ & mIoU & nIoU & $P_d$ & $F_a$ \\
		\hline
		Top-Hat \cite{zhu2019infrared} & 7.143 & 18.27 & 79.84 & 1012 & 20.72 & 28.98 & 78.41 & 166.7 & 10.06 & 7.74 & 75.11 & 1432 \\
		WSLCM \cite{han2020infrared}  & 1.158 & 6.835 & 77.95 & 5446 & 2.283 & 3.865 & 56.82 & 1309 & 3.452 & 0.678 & 72.44 & 6619 \\
		IPI \cite{gao2013infrared} & 25.67 & 50.17 & 84.63 & 16.67 & 17.76 & 15.42 & 74.49 & 41.23 & 27.92 & 20.46 & 81.37 & 16.18 \\
		ACM \cite{dai2021asymmetric}  & 68.93 & 69.18 & 91.63 & 15.23 & 61.12 & 64.40 & 94.18 & 34.61 & 59.23 & 57.03 & 93.27 & 65.28 \\
		ALCNet \cite{dai2021attentional}  & 70.83 & 71.05 & 94.30 & 36.15 & 64.74 & 67.20 & 94.18 & 34.61 & 60.60 & 57.14 & 92.98 & 58.80 \\
		RDIAN \cite{sun2023receptive}  & 68.72 & 75.39 & 93.54 & 43.29 & 76.28 & 79.14 & 95.77 & 34.56 & 56.45 & 59.72 & 88.55 & 26.63 \\
		ISTDU-Net \cite{Hou2022ISTDU-Net} & 75.52 & 79.73 & 96.58 & 14.54 & 89.55 & 90.48 & 97.67 & 13.44 & 66.36 & 63.86 & 93.60 & 53.10 \\
		IAANet \cite{wang2022interior} & 74.22 & 75.58 & 93.53 & 22.70 & 90.22 & 92.04 & 97.26 & 8.32 & 66.25 & 65.77 & 93.15 & 14.20 \\
		DNANet \cite{li2022dense} & 75.80 & 79.20 & 95.82 & \textbf{8.78} & 88.19 & 88.58 & \textbf{98.83} & 9.00 & 65.90 & 66.38 & 90.91 & \textbf{12.24} \\
		UIU-Net \cite{wu2022uiu} & 75.75 & 71.50 & 95.82 & 14.13 & 93.48 & 93.89 & 98.31 & 7.79 & 66.15 & 66.66 & 93.98 & 22.07 \\
		\textbf{MSCA-Net(Ours)} & \textbf{78.43} & \textbf{79.91} & \textbf{96.96} & 16.80 & \textbf{94.56} & \textbf{94.36} & 98.52 & \textbf{5.86} & \textbf{67.08} & \textbf{67.15} & \textbf{94.12} & 17.02 \\
		\hline
	\end{tabular}
\end{table*}
\begin{table}[t]
	\caption{Performance comparison with SOTA methods on IRSTD-1K in terms of parameters(M), Flops(G), inference time(s), and memory usage(M).}
	\label{params}
	\centering
	\renewcommand{\arraystretch}{1.3}
	\setlength\tabcolsep{1.6mm}
	\begin{tabular}{cccccc}
		\hline \hline
		Method   & Params & Flops & Inference & Memory\\ \hline
		ACM \cite{dai2021asymmetric}  & 0.398  & 0.402 & 0.029 & 897     \\
		RDIAN \cite{sun2023receptive} & 0.216  & 3.718 & 0.035 & 3331   \\
		ISTDU-Net \cite{Hou2022ISTDU-Net} & 2.751  & 7.944 & 0.039 & 10717    \\
		UIU-Net \cite{wu2022uiu} & 50.540  & 54.425 & 0.060 & 14747     \\
		DNA-Net \cite{li2022dense} & 4.696  & 14.261 & 0.050 & 9869    \\
		SCTransNet \cite{yuan2024sctransnet}  & 11.19  & 20.24 & 0.067 & 7077   \\
		\textbf{MSCA-Net(Ours)}  & 9.85  & 11.79 & 0.020 & 11031     \\
		\hline
	\end{tabular}
\end{table}
\newline \indent The experimental results demonstrate that model-driven approaches generally perform worse than data-driven methods do, particularly with a notable gap in terms of the IoU. This is attributed to the fact that model-driven approaches emphasize the approximate localization of targets rather than achieving accurate delineation of their shapes. However, in complex and dynamic backgrounds, model-driven methods often overlook small targets and their boundaries, leading to poor performance in small target detection tasks.
\newline \indent Table~\ref{params} presents a comprehensive comparison between MSCA-Net and SOTA methods on the IRSTD-1K datasets, covering model size, GFLOPs, inference time for each image, and GPU memory usage. Compared to SCTransNet, MSCA-Net reduces the number of parameters by 11.97\% and FLOPs by 41.74\%. Notably, despite a 55.87\% increase in GPU memory consumption, the inference speed per image is improved by a factor of 3.35, demonstrating a favorable trade-off between processing efficiency and memory usage.
\newline \indent Our proposed method outperforms other approaches across all three datasets, confirming its effectiveness and advantages in the IRSTD task. Compared to other methods, our MSCA-Net model is more efficient in multi-scale feature extraction, enhancing the ability to capture features. Additionally, our approach has a clear advantage in extracting contextual information and achieving deep interaction between global and local features. 
Although DNA-Net showed lower false-positive rates ($F_a$) on NUAA-SIRST and IRSTD-1K datasets, our approach significantly outperformed DNA-Net in detection accuracy and segmentation accuracy ($P_d$ improved by 1.2-3.5\% and IoU improved by 1.8-3.5\%). For the NUDT-SIRST dataset, although the detection probability ($P_d$) of DNA-Net is only 0.2\% higher, our method reduces the false alarm rate ($F_a$) to 1.5 times, achieving better detection reliability. This performance balance across data sets shows that our method is more robust between accuracy and false alarm suppression, and is especially suitable for infrared small target detection in complex background.
\newline \indent Figure \ref{fig:ROC} presents the ROC curves of several competitive algorithms. The ROC curve of the MSCA-Net algorithm outperforms those of the other methods. By appropriately selecting the segmentation threshold, MSCA-Net achieves the highest detection accuracy on the NUAA-SIRST and IRSTD-1K datasets, while maintaining the lowest false positives on the NUDT-SIRST and IRSTD-1K datasets.

\subsection{Ablation study}   
\label{sec:Ablation}
In this part, we validate the effectiveness of the MSCA-Net model through U-Net baseline experiments. Table \ref{tab: ablation} shows the ablation study results for the proposed MSCA-Net model. We progressively added the RBs (residual blocks), the DS (Deep Supervision), the MSEDA, PCBAM, and CA modules to the baseline model. The experiment was conducted on the NUAA-SIRST dataset.
\begin{table*}[htbp] 
	\small
	\centering      
	\caption{Ablation results of MSCA-Net}
	\label{tab: ablation} 
	\resizebox{\linewidth}{!}{
		\begin{tabular}{@{}cccccccccc@{}} 
			\toprule
			U-Net & +RBs & +DS & MSEDA & PCBAM & CAB & mIoU(\%) & nIoU(\%) & $P_d$(\%) & $F_a$($10^{-6}$)\\  
			\midrule
			\cmark & \xmark & \xmark & \xmark & \xmark & \xmark & 75.27 & 76.32 & 94.57 & 31.83  \\ 
			\cmark& \cmark & \xmark &\xmark & \xmark & \xmark & 76.01 & 77.54 & 94.78 & 28.63 \\ 
			\cmark& \cmark & \cmark &\xmark & \xmark & \xmark & 76.32 & 78.03 & 95.06 & 26.72 \\ 
			\cmark& \cmark & \cmark &\cmark & \xmark & \xmark & 77.31 & 78.94 & 95.44 & 24.03 \\ 
			\cmark& \cmark& \cmark& \cmark & \cmark & \xmark & 77.64 & 79.51 & 95.93 & 20.92 \\ 
			\cmark& \cmark & \cmark &\cmark & \cmark & \cmark & 78.43 & 79.91 & 96.96 & 16.80 \\ 
			\bottomrule
		\end{tabular}
	}
\end{table*}
\newline \indent The experimental results show that each module brings noticeable improvements. It is worth emphasizing that the incorporation of MSEDA, PCBAM, and CAB modules yields modest improvements in mIoU, nIoU, and $P_d$, with increases ranging from 0.40\% to 1.29\%. Meanwhile, the $F_a$ metric shows significant reductions of 10.1\%, 12.9\%, and 19.7\%, respectively. These results further demonstrate the effectiveness and robustness of the proposed modules in enhancing detection accuracy and suppressing false alarms.To further analyze their contributions, we examine the roles of the individual components in detail. 
\newline \indent The MSEDA module enhances the cross-level representation ability of small targets through a multi-scale feature fusion strategy, whereas the PCBAM module effectively focuses on key regions via a channel-space-location attention collaborative mechanism. CAB strengthens the integration of essential feature representations by emphasizing key channels through adaptive weighting. When these three components work together, they form a progressive optimization system involving scale adaptation, spatial localization, and channel enhancement. This synergy generates collaborative effects across three critical stages, i.e., feature extraction, attention selection, and feature enhancement, ultimately constructing a multi-level feature expression network with strong discriminative power for infrared small targets.
\newline \indent To further evaluate the effectiveness of the MSEDA, PCBAM, and CAB modules, comparative experiments are conducted against several SOTA counterparts for each module individually.
\newline \indent \textbf{(1) Impact of the MSEDA module:} The MSEDA component is developed to facilitate the extraction of features across multiple scales, thereby enhancing the model's ability to capture both coarse and fine-grained information. To confirm its functionality, we compared the MSEDA module with two other multi-scale modules, SCTB and ASSA. The experimental results, presented in Table \ref{tab: MSEDA-Comparative}, demonstrate that MSEDA outperforms both SCTB and the ASSA in improving the feature representation of small targets, leading to a significant increase in detection accuracy.
\begin{table}[htbp] 
	\small
	\centering      
	\caption{Comparative experiment of the MSEDA module in MSCA-Net}
	\label{tab: MSEDA-Comparative} 
	\resizebox{\linewidth}{!}{
		\begin{tabular}{@{}ccccccc@{}} 
			\toprule
			Module name & mIoU(\%) & nIoU(\%) & $P_d$(\%) & $F_a$($10^{-6}$)\\  
			\midrule
			SCTB \cite{yuan2024sctransnet} & 75.53 & 76.47 & 94.88 & 30.22  \\ 
			ASSA \cite{zhou2024adapt} & 76.07 & 76.19 & 95.03 & 24.71 \\ 
			\textbf{MSEDA} & \textbf{77.31} & \textbf{78.94} & \textbf{95.44} & \textbf{24.03} \\ 
			\bottomrule
		\end{tabular}
	}
\end{table}
\newline \indent \textbf{(2) Impact of the PCBAM module:} To validate its effectiveness, we compared it with two other attention modules, CFN and CBAM. The empirical findings, presented in Table \ref{tab: PCBAM-Comparative}, indicate that the PCBAM module outperforms the other two modules in constructing contextual dependencies and capturing the local feature representations, thereby effectively enhancing the detection precision of the model.
\begin{table}[htbp] 
	\small
	\centering      
	\caption{Comparative experiment of the PCBAM module in MSCA-Net}
	\label{tab: PCBAM-Comparative} 
	\resizebox{\linewidth}{!}{
		\begin{tabular}{@{}ccccccc@{}} 
			\toprule
			Module name & mIoU(\%) & nIoU(\%) & $P_d$(\%) & $F_a$($10^{-6}$)\\  
			\midrule
			CFN \cite{yuan2024sctransnet} & 76.98 & 79.03 & 95.08 & 23.43  \\ 
			CBAM \cite{woo2018cbam} & 77.02 & 78.88 & 94.93 & 23.11 \\ 
			\textbf{PCBAM} & \textbf{77.64} & \textbf{79.51} & \textbf{95.93} & \textbf{20.92} \\ 
			\bottomrule
		\end{tabular}
	}
\end{table}
\newline \indent \textbf{(3) Impact of the CAB module.} To verify its effectiveness, we compared it with two other feature fusion modules, the CCA and the FAM. The experimental results, shown in Table \ref{tab: cab-Comparative}, demonstrate that CAB outperforms the other two modules in integrating essential representations by adaptively emphasizing key channels, strengthening informative content, and mitigating redundant interference, thus improving the model’s segmentation precision.
\begin{table}[htbp] 
	\small
	\centering      
	\caption{Comparative experiments of CAB modules in MSCA-Net}
	\label{tab: cab-Comparative} 
	\resizebox{\linewidth}{!}{
		\begin{tabular}{@{}ccccccc@{}} 
			\toprule
			Module name & mIoU(\%) & nIoU(\%) & $P_d$(\%) & $F_a$($10^{-6}$)\\  
			\midrule
			CCA \cite{yuan2024sctransnet} & 77.83 & 78.53 & 95.46 & 19.83  \\ 
			FAM \cite{ren2024ninth} & 78.01 & 78.94 & 96.24 & 19.24 \\ 
			\textbf{CAB} & \textbf{78.43} & \textbf{79.91} & \textbf{96.96} & \textbf{16.80} \\ 
			\bottomrule
		\end{tabular}
	}
\end{table}
\begin{figure*}[htbp]
	\centering
	\includegraphics[width=.95\textwidth]{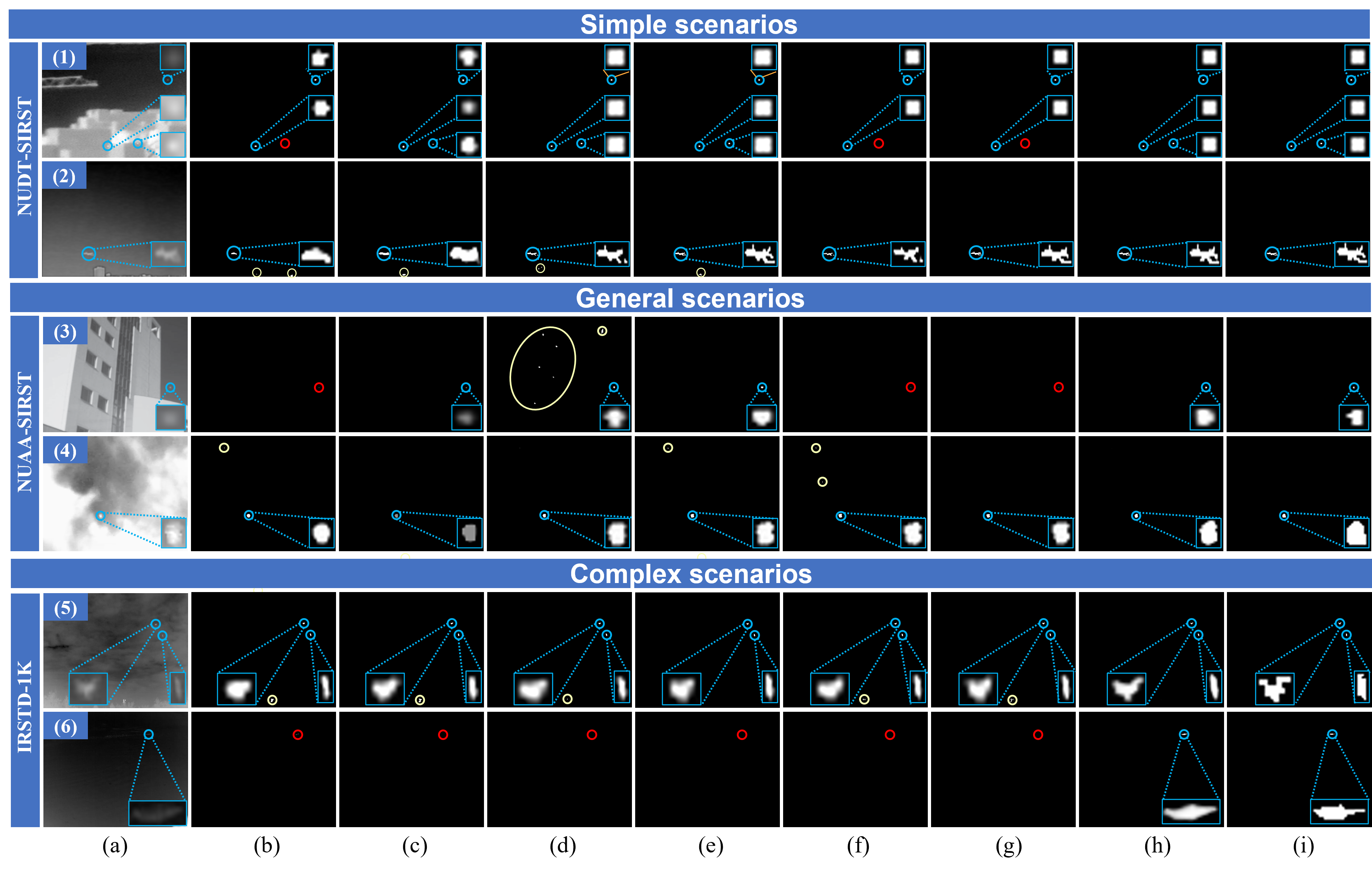}
	\caption{Visualization of the saliency maps were obtained by various IRSTD methods. The circles in blue, red, and yellow represent correctly detected targets, missed detections, and false alarms, respectively. (a) Input. (b) ACM. (c) ALCNet. (d)RDIAN. (e) DNA-Net. (f) ISTDU-Net. (g) UIU-Net. (h) MSCA-Net. (i) GT.} 
	\label{fig:visiual-images}
\end{figure*}
\subsection{Visual results}   
\label{sec:Visual}
\begin{figure*}[htbp]
	\centering
	\includegraphics[width=.95\textwidth]{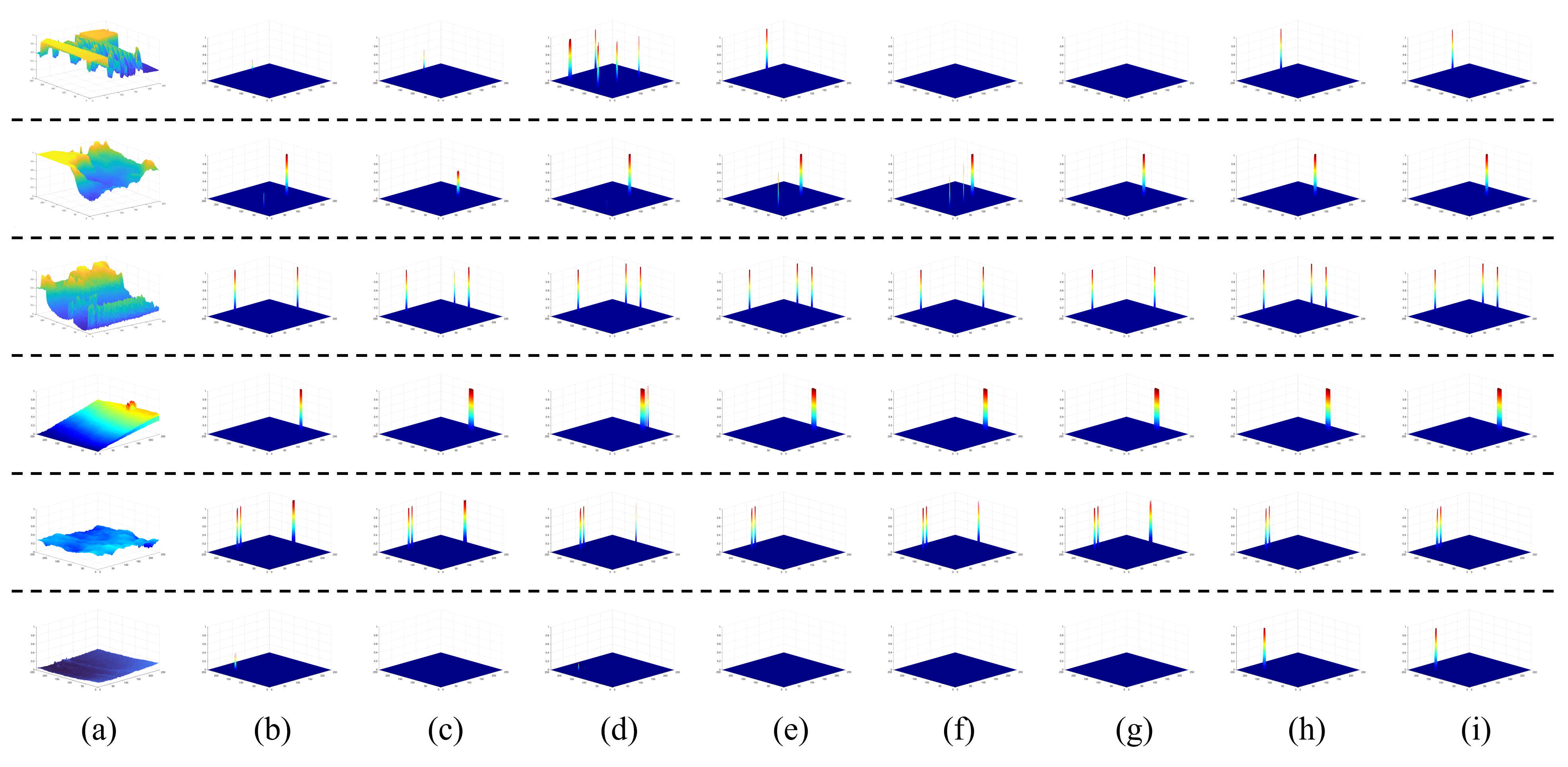}
	\caption{3-D visualization of the saliency maps of different methods on six test images. (a) Input. (b) ACM. (c) ALCNet. (d)RDIAN. (e) DNA-Net. (f) ISTDU-Net. (g) UIU-Net. (h) MSCA-Net. (i) GT.} 
	\label{fig:salientmaps}
\end{figure*}

Using three publicly available IRSTD datasets, we qualitatively compare seven representative IRSTD algorithms. The proposed method demonstrates marked improvements in reducing missed and false detections under complex scenes, including targets partially obscured by haze in the sky, embedded within bright building structures, or nearly merged with dark backgrounds ect. These advantages stem from the combined effect of multi-scale feature extraction and a position–channel–spatial attention mechanism, which jointly enhances edge response and suppresses background interference. As a result, our method achieves clearer target delineation and significantly fewer false detections in diverse and cluttered environments.
\newline \indent As shown in Figures \ref{fig:visiual-images} and \ref{fig:salientmaps}, we selected six images with complex backgrounds and considerable interference for testing. The blue boxes represent true positives, the yellow boxes denote false positives, and the red boxes highlight missed detections. The experimental results indicate that, compared with other algorithms, our method successfully detects every target, whereas other methods result in some missed and false detections. For instance, as shown in Figure \ref{fig:visiual-images} (5), all methods, except for our method and DNA-Net, produced false alarms. This issue can be attributed to these methods relying solely on local contrast information and lacking the ability to model long-range dependencies in the image, making them prone to false positives in complex backgrounds. As shown in Figure \ref{fig:visiual-images} (6), our method successfully segmented and localized the target, whereas other deep learning methods failed to detect it. Figure \ref{fig:salientmaps} displays a 3D representation of the saliency maps produced by various methods across the six test images. This is consistent with the visual results shown in Figure \ref{fig:visiual-images}.
\section{Conclusion}   
\label{sec:conclusion}
This study proposes a multi-scale perception-based infrared small target detection algorithm, MSCA-Net, which is based on the U-Net architecture and employs a triple-coordinated optimization mechanism to overcome performance bottlenecks. Specifically, in the encoder stage, a Multiscale Enhanced Dilated Attention is introduced, combined with a cascaded dilated convolution module to adaptively aggregate multi-scale information, thus strengthening the model’s ability to extract discriminative features. Moreover, a Positional Convolutional Block Attention is added to strengthen the feature response of key areas through the space-channel-position triple attention mechanism, promoting the coordinated representation optimization of local details and global semantics. In the decoder stage, a Channel Aggregation Feature Fusion Block is further incorporated to consolidate informative representations by adaptively highlighting pivotal channels. This facilitates effective feature propagation and enhances the model’s robustness under complex. The experimental outcomes reveal that MSCA-Net delivers remarkable detection capabilities across three benchmark datasets—NUAA-SIRST, NUDT-SIRST, and IRSTD-1K—where quantitative analyses robustly affirm its efficacy and advantages.

\bibliographystyle{cas-model2-names}
\bibliography{cas-refs.bib}

\end{document}